\documentclass[runningheads]{llncs}
\usepackage[T1]{fontenc}
\usepackage{colortbl}
\usepackage{xcolor}
\usepackage{caption}
\usepackage{amsmath}
\usepackage{placeins}
\usepackage{makeidx}
\usepackage{subcaption}
\usepackage{url}
\usepackage{hyperref}
\usepackage{algorithm}
\usepackage{algpseudocode}

\usepackage{graphicx}
\begin{document}
\title{Force-Driven Validation for Collaborative Robotics in Automated Avionics Testing}
\titlerunning{Force-Driven Validation in Avionics Testing}
%
\author{Pietro Dardano\inst{1}, Paolo Rocco\inst{2} \and David Frisini\inst{3}}
\authorrunning{Pietro Dardano et al.}

\index{Dardano, P.}
\index{Rocco, P.}
\index{Frisini, D.}

\institute{Dipartimento di Elettronica, Informazione e Bioingegneria, 
Politecnico di Milano, Piazza L. da Vinci 32, 20133, Milano, MI, Italy and  
TXT E-TECH s.r.l., Via Milano 150, 20093 Cologno Monzese, MI, Italy. \email{pietro1.dardano@mail.polimi.it}
\and
Dipartimento di Elettronica, Informazione e Bioingegneria, 
Politecnico di Milano, Piazza L. da Vinci 32, 20133, Milano, MI, Italy.\\
\and 
TXT E-TECH s.r.l., Via Milano 150, 20093, Cologno Monzese, MI, Italy}
\maketitle              
\begin{abstract}
  ARTO is a project combining collaborative robots (cobots) and Artificial Intelligence (AI)
  to automate functional test procedures for civilian and military
  aircraft certification. This paper proposes a Deep Learning (DL)
  and eXplainable AI (XAI) approach, equipping ARTO with interaction analysis
  capabilities to verify and validate the operations on cockpit
  components. During these interactions, forces, torques, and end-effector
  poses are recorded and preprocessed to filter disturbances caused by
  low-performance force controllers and embedded Force Torque Sensors
  (FTS). Convolutional Neural Networks (CNNs) then classify the cobot
  actions as \textbf{Success} or \textbf{Fail}, while also identifying and reporting the causes of
  failure. To improve interpretability, Grad-CAM, an XAI technique for
  visual explanations, is integrated to provide insights into the model’s
  decision-making process. This approach enhances the
  reliability and trustworthiness of the automated testing
  system, facilitating the diagnosis and rectification of errors that may
  arise during testing.

\end{abstract}

\vspace{-0.8cm}
\begin{center}
  \url{https://github.com/pietrodardano/ForceDriven_Arto}
\end{center}
\vspace{-0.6cm}
\section{Introduction}
Ensuring the safety and reliability of civilian and military aircraft 
requires extensive certification procedures, involving 
a suite of functional and non-functional tests, many of which are 
still performed manually by skilled operators. While manual testing 
remains a standard practice, it is inherently prone to human error 
and inconsistencies. This introduces potential risks and inefficiencies 
into a process where reliability is paramount. The need for a robust, 
repeatable, cost-effective and efficient testing solution has driven the development 
of new methodologies leveraging robotics and AI.

Automated Robotics for Testing Optimisation (ARTO) addresses these 
challenges by integrating cobots, AI and advanced sensing capabilities. 
The system automates cockpit test procedures, enabling a cobot to interact with various 
cockpit components such as touchscreens, buttons, levers, and knobs.

\begin{figure}[h]
    \centering
    \begin{minipage}{0.32\textwidth}
        \centering
        \includegraphics[width=\textwidth]{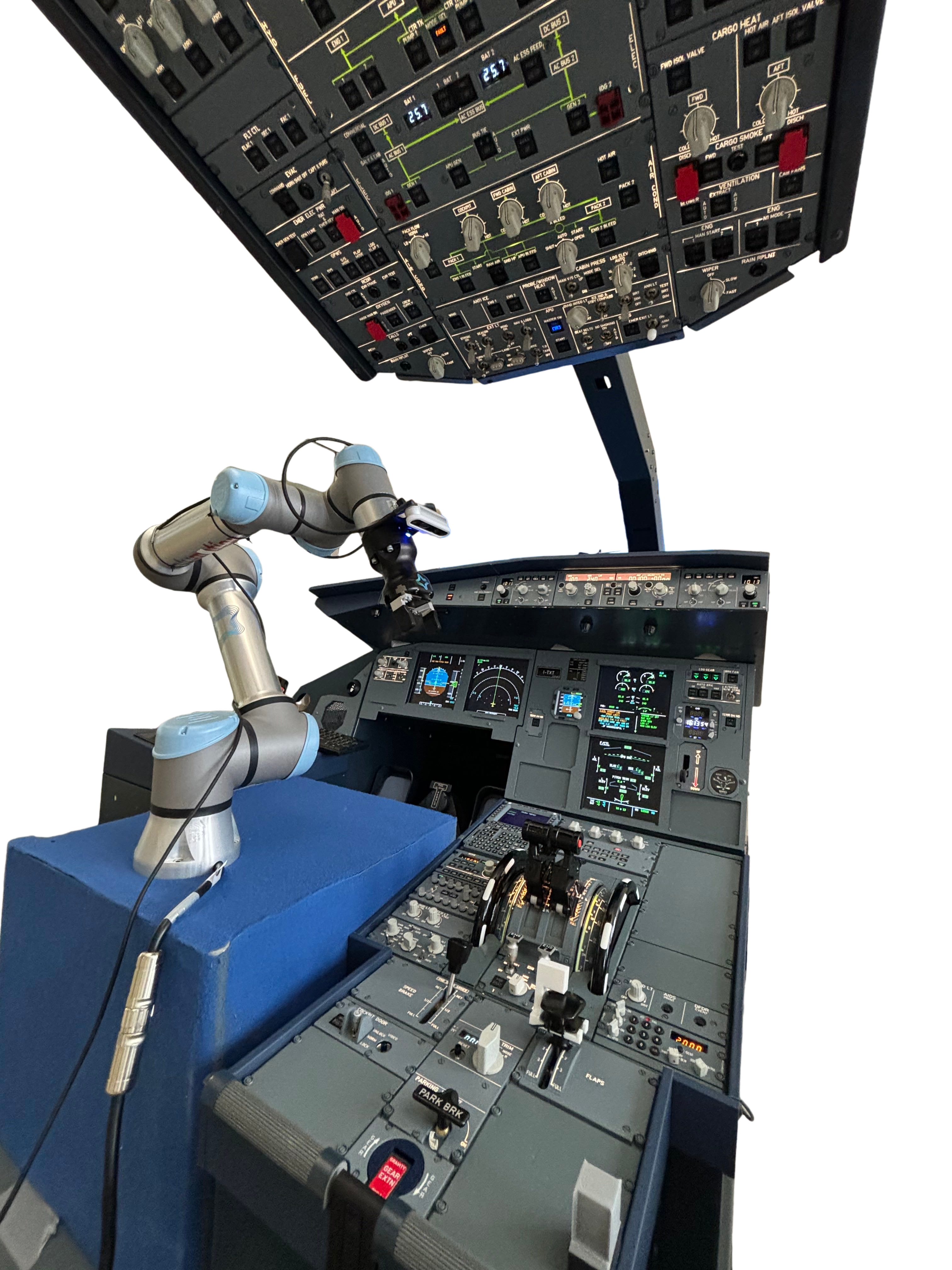}
        \caption{ARTO setup}
        \label{fig:setup_flap_button_a}
    \end{minipage}%
    \hspace{0.006\textwidth}
    \begin{minipage}{0.32\textwidth}
        \centering
        \includegraphics[width=1.333\textwidth, angle=-90]{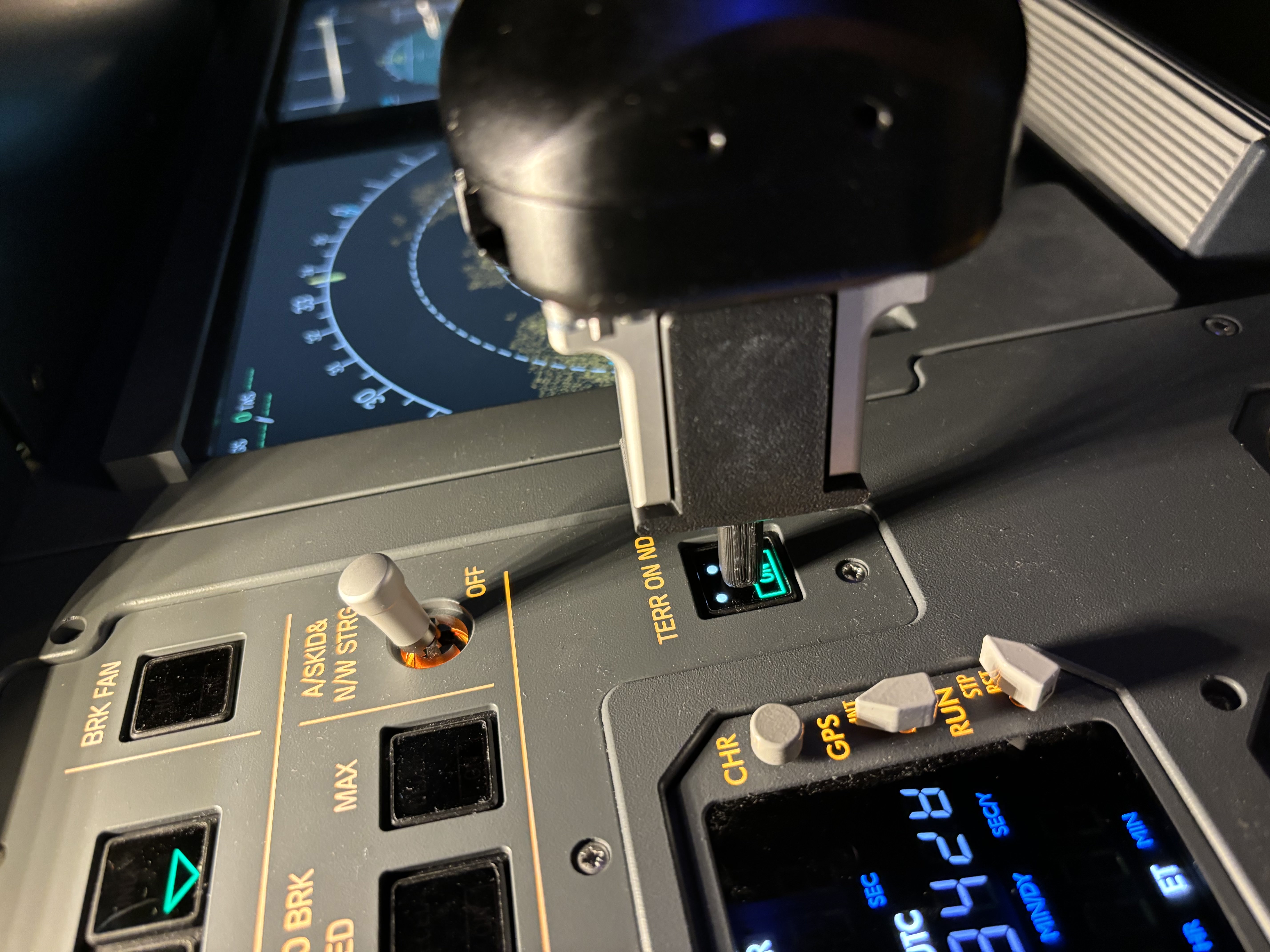}
        \caption{Button pressing}
        \label{fig:setup_flap_button_b}
    \end{minipage}%
    \hspace{0.006\textwidth}
    \begin{minipage}{0.32\textwidth}
        \centering
        \includegraphics[width=1.333\textwidth, angle=-90]{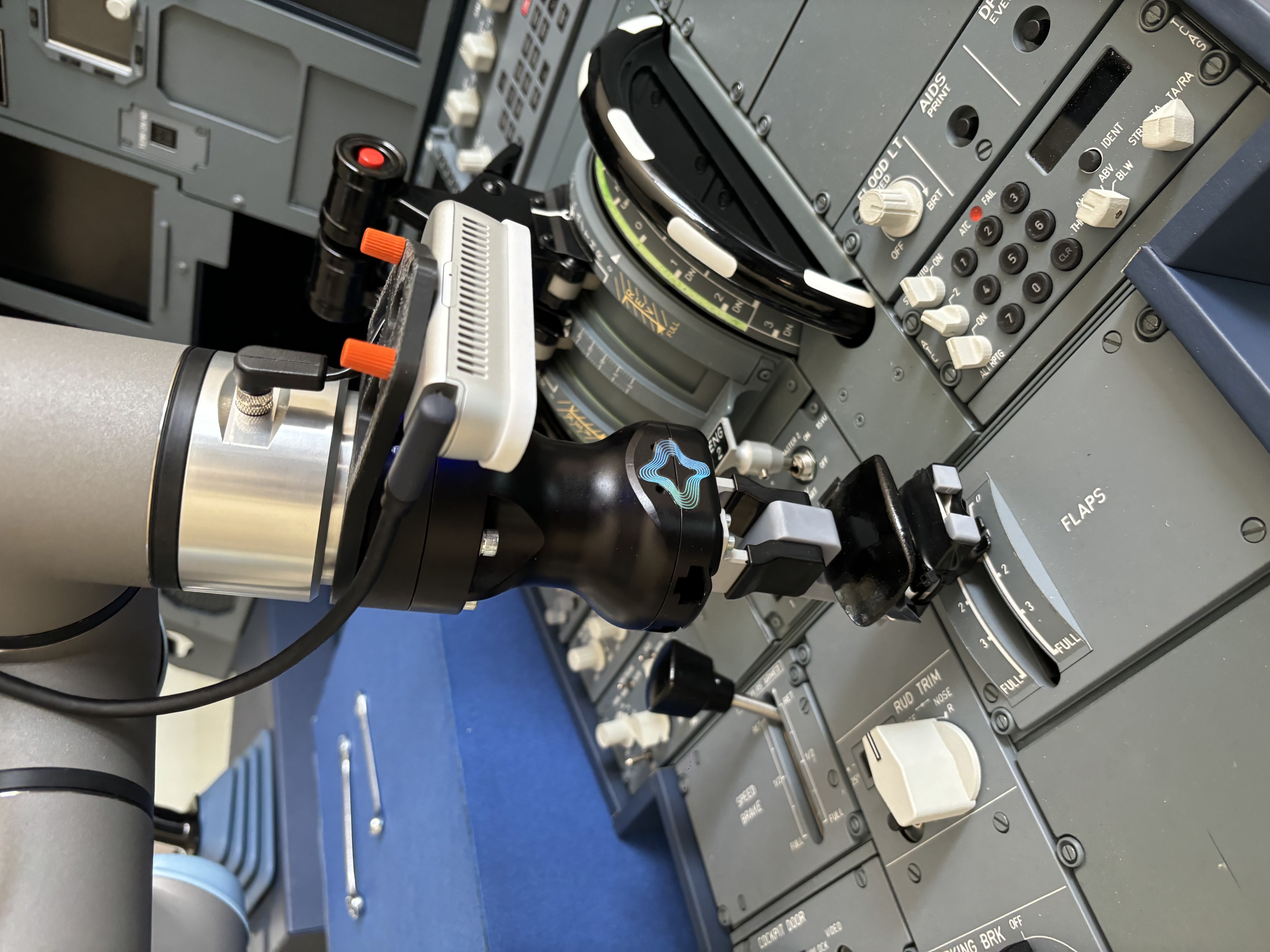}
        \caption{FLAP actuation}
        \label{fig:setup_flap_button_c}
    \end{minipage}
    \vspace{-0.52cm}
\end{figure}
The cobot executes test procedures based on predefined sequences or 
automatically extracted instructions from test procedure documents.
A trajectory generation algorithm, enables the robot to navigate 
safely within the cockpit environment, avoiding collisions while reaching designated approach points.
For the component's actuation we leverage the UR5e's
hybrid force-position controller and the embedded FTS however,
due to the low resolution and accuracy of the embedded sensor, 
preprocessing techniques are necessary to ensure high-fidelity interaction data.

Custom 3D-printed tools serve as intermediary attachments, 
enabling the robot to manipulate objects more easily 
and protecting the cockpit components in case of unexpected maneuvers.
Data acquisition involves capturing multi-modal sensor readings, including: 
force, torque, joint angles, and end-effector poses.
To ensure robust analysis, an energy-based transient isolation algorithm 
filters out disturbances generated by low-performance controller and sensor inaccuracies.
This refined dataset is used to train DL models for the 
classification and validation of the performed actions.
The work develops and evaluates CNNs tailored for one-dimensional (1D) signals 
and their two-dimensional (2D) representations obtained using the Continuous Wavelet Transform (CWT).
Hybrid-CNNs combines and leverages both representations, are also developed.

These models classify cobot actions as \textbf{Success} or \textbf{Fail}, 
while also identifying failure causes. 
To improve interpretability and foster trust in AI-driven decisions, 
XAI techniques particularly Gradient-weighted Class Activation Mapping (Grad-CAM) are integrated. 
This method highlights regions of the data, as a visual explanation, that most influence the CNN’s 
predictions, providing valuable insights for operators and system integrators. 

The developed models are integrated into the ARTO's framework using Robot Operating System 2 (ROS2), 
ensuring real-time processing and seamless communication between system components.
This study demonstrates the feasibility of an AI-driven,
wrench-based validation framework for avionics testing, 
contributing to advancements in aerospace testing methodologies by 
improving efficiency, reliability, and safety.




\section{Related Work}\vspace{-0.1cm}
Avionics systems must adhere to safety standards like ARP4761 and DO-178C,
which require extensive testing traditionally performed by engineers, 
resulting in high costs and potential human errors. 
By automating this process, multiple tests can be 
conducted in parallel, helping to manage costs effectively.
\vspace{-0.2cm}

\subsection{Automated Testing in Avionics}
The ARTO concept, exploring robotics and computer vision
for automated testing, was introduced in 2022~\cite{frisini2022technology}.
Initially at Technology Readiness Level (TRL) 1-2, it analyzed basic principles
and formulated the technology. By 2023, the first implementation
of ARTO reached TRL 3-4~\cite{frisini2023evaluation},
employing an ABB GoFa 1500 cobot and a 3D-printed rigid end-effector for 
touchscreen usage, button pressing, and landing gear operation. 
However, it lacked force control and contact sensing, 
relying solely on computer vision to detect action outcomes. 
Moreover, a fixed database of predefined actions and poses constrained the system’s adaptability.\\
Alternative approaches such as ROSSI (Airtificial Group) 
and ROCCET (a Lufthansa-backed startup), are still in early R\&D phases and not yet applied to full aircraft cockpits~\cite{rossi2022,roccet2019}. 
Unlike ARTO, which operates on real aircraft or certified simulators, 
these projects focus on subsystems or unofficial ones.  
With recent advancements and latest validation tests ARTO now reaches TRL 5-6.
\vspace{-0.2cm}

\subsection{Wrench analysis and Success Detection}
The closed-source nature of many robot controllers restricts 
the ability to customize and fine-tune the provided low-level force control,
necessitating the development of external signal-processing 
techniques to enhance functionality and add features.  
Early approaches employed Machine Learning (ML): 
using SVMs, LSTMs, and MLPs for object recognition, contact detection, 
and actions classification in robot-environment and human-robot interactions, 
showcasing the potential of ML-based methods~\cite{eiband2021identification,zhang2022finger,castro2023classification,riffo2022object,stolt2015detection}.
Tsujii et.al. demonstrated the effectiveness of Mel-Frequency Cepstral Coefficients with Time-Delay
Neural Networks, a form of 1D-CNN, for recognizing click responses
\cite{tsuji2021contact}. Building on these foundations and on the work of Stolt et.al., 
the use of CNNs on force signals for object recognition, 
inspired the DL architectures developed in this research
\cite{stolt2015detection}. 

To improve detection accuracy in our developed models, 
an energy-based transient detection algorithm isolates 
informative signal features~\cite{waghmare2012transient}, during the preprocessing phase. 
CNNs have demonstrated strong performance in signal classification, 
with 1D CNNs achieving competitive results after fine-tuning~\cite{castro2023classification,tsuji2021contact,stathatos2020cnn}.  
In parallel, Continuous Wavelet Transforms (CWT) provide 2D signal representations, 
capturing both frequency and time-domain information~\cite{russell2016jean,torrence1998practical}. 
These 2D scaleograms can be processed alongside 1D signals 
within a multi-branch CNN architecture, enhancing overall classification accuracy~\cite{lee2021vision,chen2018online,sun2021hybrid}.

Inspired by previous research~\cite{zhou2015,shi2023}, we developed a 
multi-branch Grad-CAM approach applicable to single- and multi-input models.  
Initially leveraging GlobalAveragePooling, we found that GlobalMaxPooling 
better highlights discriminative regions, aligning with our goals.  
\section{Setup and Data Preprocessing}
ARTO now includes an Airbus A320 Single-Seat simulator and a 6-DOF Universal Robots UR5e cobot, 
positioned at the pilot's station. The cobot is equipped with a RobotiQ Hand-E gripper with rubber pads, 
enabling it to interact with cockpit components (Fig.~\ref{fig:setup_flap_button_a}).  
Designed 3D-printed tools, including the finger and flap attachments,  
are shown in Fig.~\ref{fig:setup_flap_button_b} and Fig.~\ref{fig:setup_flap_button_c} during operation. 
The system performs testing on the following cockpit components: 
\textbf{Buttons}, \textbf{Switches}, \textbf{Knobs}, \textbf{FLAP}, \textbf{Landing Gear (LDG)}, and \textbf{Speed-Brake (S-Brake)}.

Actions are executed leveraging the UR5e’s hybrid force-position controller, 
accessed via the UR\_RTDE control interface function \texttt{Force\_Mode()}. 
During execution, the cobot records forces, torques, 
and Tool Center Point (TCP) poses in the robot's base frame. 
The embedded Force Torque Sensor (FTS) samples data at 500 Hz 
but has limited resolution and precision. 

During data collection, once an action is performed and stored, 
a label is assigned via script. 
Since this is a supervised learning problem, this label serves as ground truth  
for classification and represents the action’s outcome. 
An extensive data collection process resulted in a dataset of over
4330 manually labeled actions. 
Future developments will explore semi-supervised learning,  
integrating Computer Vision together with other perception modalities.  
For button interactions, only two classes are considered: 
\texttt{Success} and \texttt{Fail}.  
For all the other components, multiple classes define  
successful transitions to different states.
Failure cases include situations such as insufficient knob rotation,  
leading to a mid-state, LDG lever not completely moved up (down), or FLAP remains locked.  

\begin{figure}[h]
    \centering
    \begin{subfigure}{0.49\textwidth}
        \centering
        \includegraphics[width=0.98\textwidth]{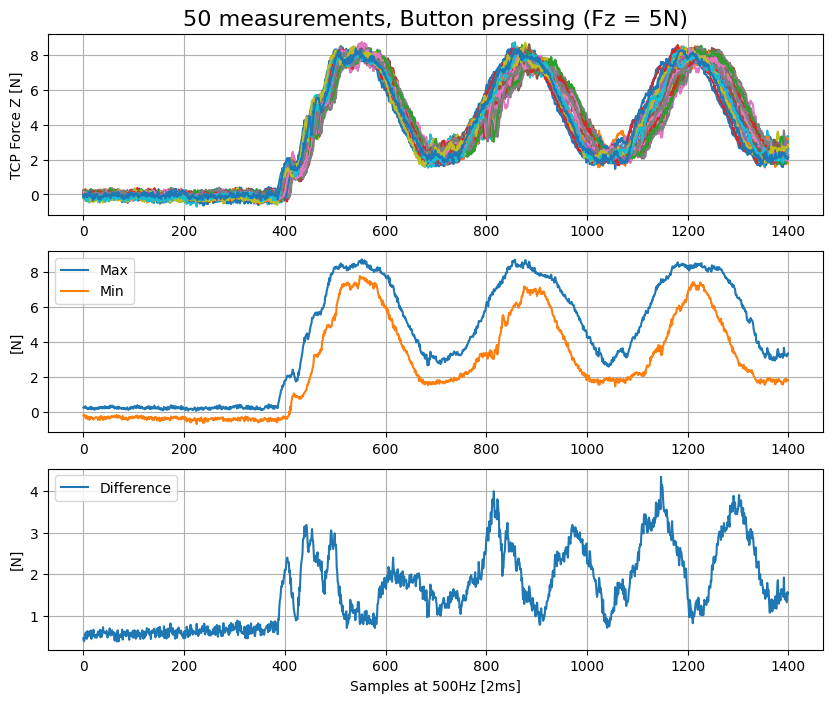}
        \caption{Oscillations and errors}
        \label{fig:oscillations}
    \end{subfigure}%
    \hspace{0.01\textwidth}
    \begin{subfigure}{0.49\textwidth}
        \centering
        \raisebox{8mm}{ 
            \includegraphics[width=\textwidth]{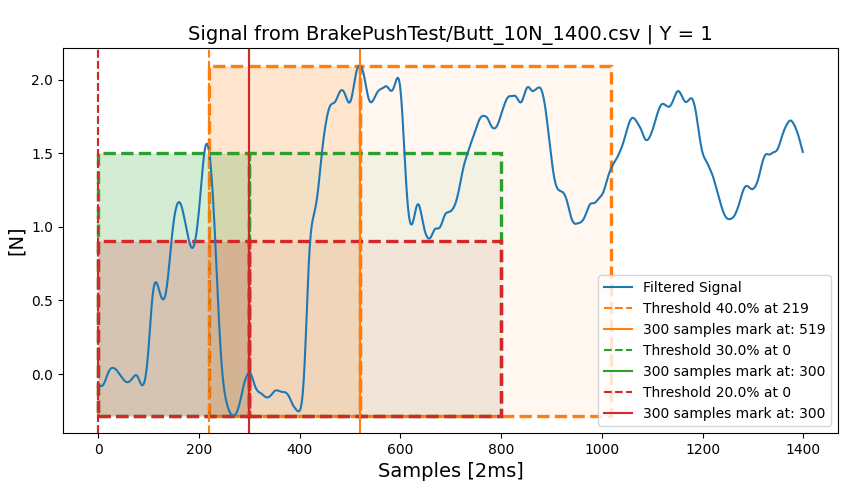}
        }
        \caption{Main transient detection}
        \label{fig:energy_distant}
    \end{subfigure}
    \caption{ (a) Measured force feedback during multiple button pressing actions, highlighting oscillations and variability across trials. (b) Signal segmentation using the developed algorithm, which isolates the most informative transient phase by detecting key energy-based thresholds and applying a fixed-length window.}
    \label{fig:oscillations_and_energy-algorithm}
\end{figure}

TCP orientation data, represented as Euler angles,  
is excluded due to the presence of potential discontinuities
which could introduce inconsistencies in the classification process.
The robot's force controller relies on these measurements,  
resulting in inconsistent force application.  
This issue is especially pronounced in pressing actions,  
where force exhibits oscillatory behavior (Fig.~\ref{fig:oscillations}).  
Another aspect to consider when using the UR5e and the UR\_RTDE library,
is that recorded forces and torques are reported in the robot's base frame rather  
\begin{minipage}[h]{0.49\textwidth}
\vspace{0.1cm}
than the TCP frame.
Based on the TCP orientation,
these values can be transformed accordingly to Eq.\ref{eq:force_torque_transformation} with
\(\mathbf{R}^\top\) being  the transposed rotation matrix, 
\end{minipage}
\hfill
\begin{minipage}[h]{0.49\textwidth}
  \begin{equation}
    \begin{bmatrix}
    \mathbf{F}_\text{tcp} \\
    \mathbf{T}_\text{tcp}
    \end{bmatrix}
    =
    \mathbf{R}^\top
    \begin{bmatrix}
    \mathbf{F}_\text{base} \\
    \mathbf{T}_\text{base}
    \end{bmatrix}
    \label{eq:force_torque_transformation}
    \end{equation}
    \vspace{0.1cm}
\end{minipage}
derived from the rotation angles \([r_x, r_y, r_z]\) in the TCP pose.
Noisy signals are filtered using a Butterworth filter  
with a cutoff frequency (CF) of 30 Hz. 
This value was determined by running CNN models 
and comparing their average classification accuracy. 

Additional preprocessing techniques include normalization, 
either using Standard or MinMax scaling,
and padding at both ends of the signals. 
For imbalanced datasets or small sample sizes, 
augmentation techniques such as: time dilation (compression),  
traslation, and white noise addition are applied.  
However, cropping, slicing, inversion, amplitude scaling,  
and permutation are avoided to prevent altering critical data features  
that, together with the earlier mentioned oscillation problem,
could lead to incorrect classifications.
To mitigate this latter issue, a main transient detection method based on multi-energy thresholds~\cite{waghmare2012transient}  
isolates the signal's most relevant segment.

\vspace{-0.2cm}
This algorithm first identifies the maximum energy value in a window of 300 samples.
Three threshold values (0.38, 0.22, 0.12 of the max energy) 
are then applied to locate key sample indices (Criteria~\ref{alg:selection_criteria}). 
These values are determined heuristically, posing a potential
risk of overfitting to this specific setup and force controller settings.
To address this, a more generalizable approach is currently under investigation.
Finally, a window of 800 samples (1.6s) is applied to extract and
isolate the informative segment of the signal,  
using the starting index identified by the thresholding process (Fig.~\ref{fig:energy_distant}).
\vspace{-0.2cm}
\begin{algorithm}[h]
    \captionsetup{labelformat=empty}
    \caption{Selection Criteria}
    \begin{tabular}{l}
        \textbf{If} the indices are within 60 samples of each other: \\
        \quad Choose the middle index. \\
        \textbf{Else if} the 38\% index is more than 200 samples away: \\
        \quad Select the 38\% index (as smaller indices may be disturbances). \\
        \textbf{Else}: \\
        \quad Choose the smallest index (12\%).
    \end{tabular}
    \label{alg:selection_criteria}
\end{algorithm}
\vspace{-0.2cm}\\
To use 2D-CNN and Hybrid CNN models,
the one-dimensional time-domain signals are transformed into 2D representations.  
This transformation is applied only to signals with high variance,  
such as force and torque, as they contain information-rich 
features by applying the CWT (Eq.~\ref{eq:dcwt}) using the Morlet wavelet~\cite{torrence1998practical} 
which is chosen for its balance between time and frequency resolution, 
making it well-suited for capturing transient of non-repetitive 
variations in signals, without assuming periodicity or stationarity.

\begin{minipage}[h]{0.46\textwidth}
  \centering
  \includegraphics[width=0.62\textwidth]{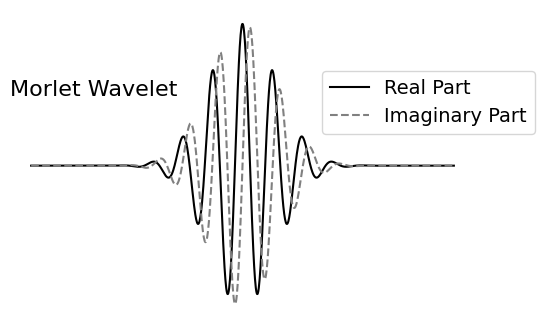}
\end{minipage}
\hfil
\begin{minipage}{0.49\textwidth}
  \vspace{-3.5mm}
  \begin{equation}
    W_x[a,b] = \sum_{n=0}^{N-1} x[n] \psi^*\left(\frac{n - b}{a}\right)
    \label{eq:dcwt}
  \end{equation}
\end{minipage}
Where \(x[n]\) is the signal, \(\psi^*\) is the complex conjugate
of the mother wavelet \(\psi\), \(a\) is the scale factor
and \(b\) is the translation one.  
The CWT produces a scaleogram \(W_x\), a 2D representation of the signal, 
where the x-axis represents time and the y-axis represents scale.  
This scaleogram is normalized into an Nx128 matrix,  
which serves as input for the 2D-CNN and Hybrid models.

Once preprocessed, the signals are stacked in depth, forming a 3D tensor.
Since different actions have varying lengths (\(N\)),  
excessive padding is avoided by creating separate datasets per action.

\vspace{-0.1cm}
\section{Classification Task}
\vspace{-0.1cm}
Four network types are developed and compared: \texttt{FF\_ANN}, \texttt{1D-CNN} for time-series,
\texttt{2D-CNN} for scaleograms and \texttt{Hybrid-CNN} combining both representations \cite{sun2021hybrid}.
Each action classification model is tailored to a specific type of robotic interaction,
such as button pressing, Knob rotation, LDG lever actuation, or FLAP movement.
For each action category, the networks process either \textbf{All} 
signals or a subset of \textbf{Selected} ones, the most informative channels, 
through Principal Component Analysis (PCA) and Max-Energy analysis,
filtering out those prone to causing misclassifications.
Manual selection can also be performed by examining the correlation matrix of the signals.
Since each model focuses on only one action category, the signals tend to exhibit similar features,
making it straightforward to group or separate them based on correlation.
These groupings inform the model’s architecture; 
highly correlated signals are processed together in the same branch, 
while less correlated or critical channels may be assigned a dedicated branch.
This design ensures that each network leverages action-specific signal properties,
enhancing classification accuracy by reducing confusion from noisy or irrelevant features prior to training.

For each network type, multiple architectures have been developed and compared  
aiming to improve both interpretability and performance.  
To ensure robust results, multi-branch architectures are employed,  
allowing signals to be processed in parallel with distinct hyperparameters.

Highly correlated signal groups are identified and assigned to specific branches,  
ensuring that structurally related signals are processed together.  
This approach improves classification accuracy by leveraging specialized feature extraction per branch.  
Kernel sizes and strides are ad hoc tuned to capture broad signal patterns.  
Kernel widths generally range from 20 to 100 samples,  
while stride values are set between 10\% and 50\% of the kernel size.  
The number of convolutional filters typically doubles at each subsequent layer as it allows to capture more complex features and hierarchies as the depth increases. 
However, in certain multi-branch configurations, filters start large,  
are reduced mid-branch for selective feature extraction,  
and then increase again toward the output.

Dropout layers of about 20\% to 40\%, combined with Max or Global Max Pooling,
reduce the parameter count, improve generalization,
and guide the model to concentrate on high-value portions of each signal.
This typically corresponds to the mid- to late-transient region,
where essential cues for success or failure are found.
Common practice is incorporating a set of Dense layers after the convolutional stacks,
gradually decreasing the number of nodes down to a minimum of 16.
This approach avoids directly flattening from the Global Max Pooling layer into the final output.
Instead, intermediate Dense layers process the extracted features,  
consistently outperforming direct connections.

Training the most complex models, such as hybrid multi-input, multi-branch architectures,
takes under one minute on an Nvidia RTX 4090 GPU with cuDNN and TensorRT acceleration,
thanks to an optimized design capped at about 8M parameters in the largest and deepest network.
Even on a lower-tier laptop equipped with an Nvidia GTX 1650Ti,
the same model trains in under three minutes, demonstrating their lightweight nature
of the developed architectures.

\vspace{-0.25cm}
\begin{figure}[h!]
    \centering
    \begin{subfigure}{0.62\textwidth}
        \centering
        \raisebox{3.5mm}{ 
            \includegraphics[width=\textwidth]{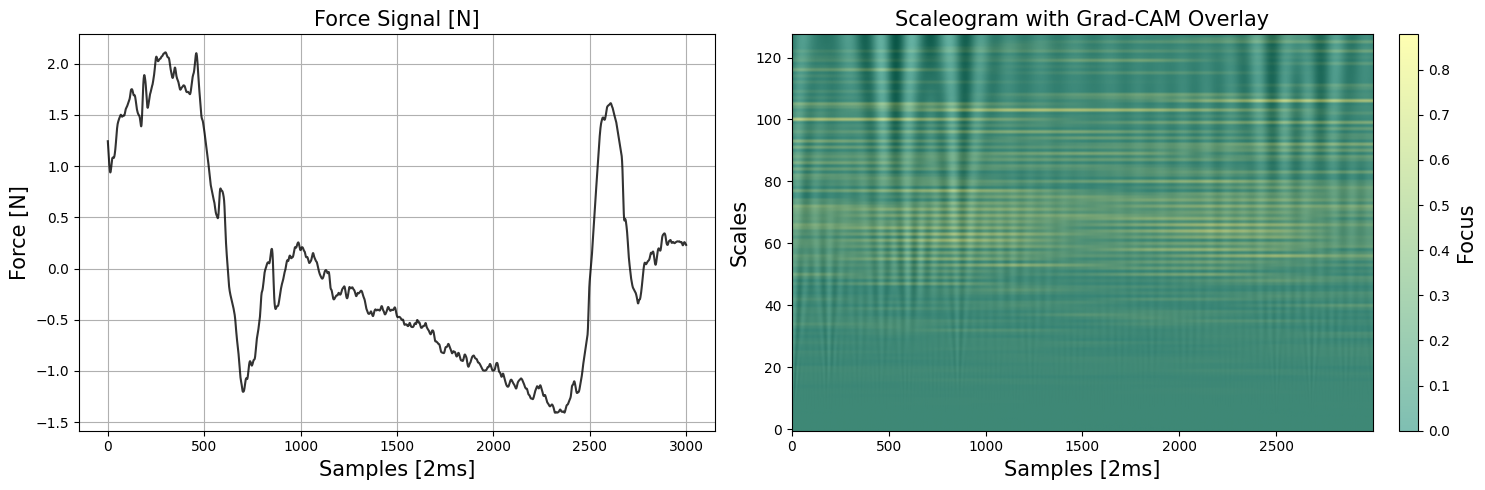}
        }
        \caption{Signal compared to its scaleogram plus Grad-CAM}
        \label{fig:2D_gradcam}
    \end{subfigure}%
    \hspace{0.01\textwidth}
    \begin{subfigure}{0.36\textwidth}
        \centering
        \includegraphics[width=0.98\textwidth]{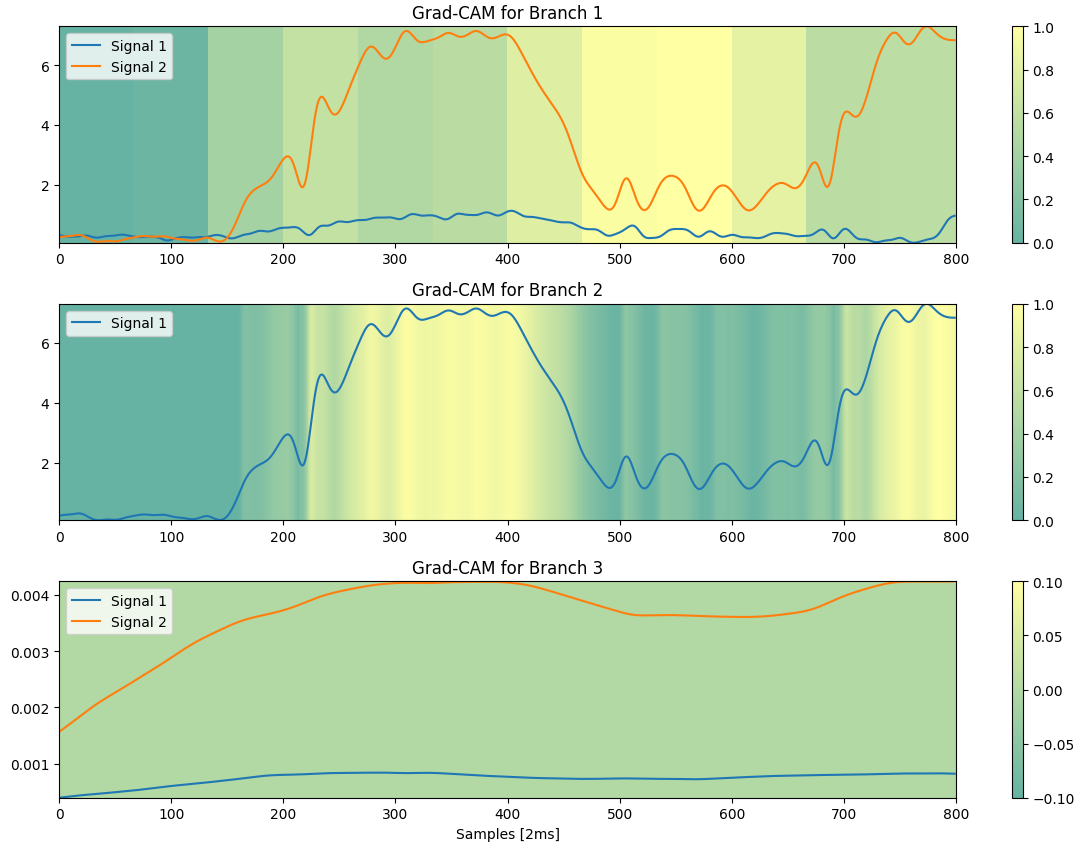}
        \caption{1D-GradCAM example}
        \label{fig:1D_gradcam_butt}
    \end{subfigure}
    \caption{Comparative analysis of 2D and 1D Grad-CAM for feature attribution. (a) The 2D variant maps activation on the wavelet-transformed signals. 
    (b) The 1D approach preserves temporal information, offering a clearer interpretation of model focus on raw force signals.}
    \label{fig:xai_comparison}
    \vspace{-0.2cm}
\end{figure}
\vspace{-0.3cm}

To elucidate the network’s decision-making process, Grad-CAM is integrated.
While its two-dimensional variant is widely used in image classification~\cite{shi2023},
applying it to wavelet-transformed scaleograms hinders interpretability,
since the original signals are abstracted through transformation.
Although the generated activation maps highlight regions of focus,
their direct interpretability is limited under this representation.

In response, one-dimensional Grad-CAM is implemented,
preserving the temporal structure and enabling verification of whether the model
attends to genuine action features or mere noise artifacts.
Unlike its two-dimensional counterpart, it does not rely on global average pooling,
thus retaining higher feature-map resolution and improving interpretability
in multi-branch architectures (Fig.~\ref{fig:xai_comparison}).
The process begins with a forward pass to obtain the model’s output,
typically a class probability. The gradients of this output with respect
to the final convolutional layer’s activations are then computed.
By applying a max-pool operation to these gradients, weight coefficients are derived,
indicating the importance of each filter. Multiplying the filter activations
by their respective weights and summing them produces a weighted feature map
across the time dimension.

This approach provides engineers and technicians,
familiar with the system and its behaviors,
an intuitive tool to diagnose model decisions,
ensuring predictions rely on meaningful features rather than disturbances.
\vspace{-0.15cm}

\section{Experiments and Results}
\vspace{-0.1cm}
\texttt{FF\_ANN} represent the simplest models  
and are the first to be tested.\\
For \texttt{Button-pressing} action classification, 
we initially tested a simple 2-layer dense model using \texttt{Not-Normalized} signals, 
achieving an F1-score of 83.5\%, compared to 76.3\% with \texttt{Standard-Scaled} signals. 
The best \texttt{FF\_ANN} model, with a depth of 5 layers, 
achieved an F1-score of 90.9\% with 2.96 million parameters using \texttt{Not-Normalized} data.
\vspace{-0.15cm}

\subsection{1D Convolutional Neural Networks}\label{subsec:1d-cnn}
\vspace{-0.1cm}
Considering again \texttt{Button-pressing} actions, the simplest 2-convolutional-layer 
model achieved an F1-score of 92.1\% with 436k parameters. 
This confirm that CNNs can achieve higher performance with less complexity. 
Deeper, multi-input multi-branch architectures further improved this result,  
achieving an F1-score of 97.8\% (Tab.~\ref{table:1D_results}).  
Notably, non-normalized signals performed better for button pressing tasks  
due to significant variability among physical buttons tested.  
Normalization, despite improving generalization,  
reduced classification accuracy by diminishing essential force-level variations,  
critical for differentiating between buttons with different required pressing forces.

Fig.~\ref{fig:1D_gradcam_butt} presents Grad-CAM visualizations applied to  
the 1D CNN model, highlighting input data from each branch.  
Since signals are not normalized, the network primarily  
focuses on the dominant force component (Fz),  
ignoring minor signals such as DeltaTCP position  
that provide minimal discriminative power for classification.
Wide regions in the heatmap reflect the use of large 
\texttt{Kernel} and \texttt{Strides} values, while narrow regions reflect smaller ones.
\begin{minipage}[t]{0.50\textwidth}
For all other action types, optimal performance 
is obtained using \textbf{Selected data} normalized by Standard Scaling, 
as reported in Tab.~\ref{table:1D_results}. 
Given the limited dataset size, a 60-20-20 (training-validation-test) split 
is chosen to provide reliable performance estimates.
If then the model achieves good performance, it is re-trained with a Train-Test split of
\end{minipage}%
\hfill
\begin{minipage}[t]{0.48\textwidth}
      \centering
      \footnotesize
      \vspace{-0.1cm}
      \begin{tabular}{|c|c|c|c|}
      \hline
      \textbf{Element } & \textbf{Scaled }  & \textbf{F1-Score } & \textbf{Params } \\ \hline
      Button  & No &  97.8\% & $\sim$1.3M \\ \hline
      Knob    & Yes &  95.2\% & $\sim$3.1M \\ \hline
      Switch  & Yes & 97.4\% & $\sim$0.5M \\ \hline
      FLAP    & Yes &  97.5\% & $\sim$1.2M \\ \hline
      LDG     & Yes &  98.5\% & $\sim$1.8M \\ \hline
      S-Brake & Yes &  98.0\% & $\sim$1.5M \\ \hline
      \end{tabular}
      \captionof{table}{1D-CNN: Results.}\label{table:1D_results}
      \vspace{0.05cm}
\end{minipage}
70-30. In addition to accuracy and F1-Score, 
confusion matrices are key factors in assessing classifier reliability.

\begin{figure}[h]
    \centering
    \subfloat[Fail: Not moved down enough]{\includegraphics[width=0.46\textwidth]{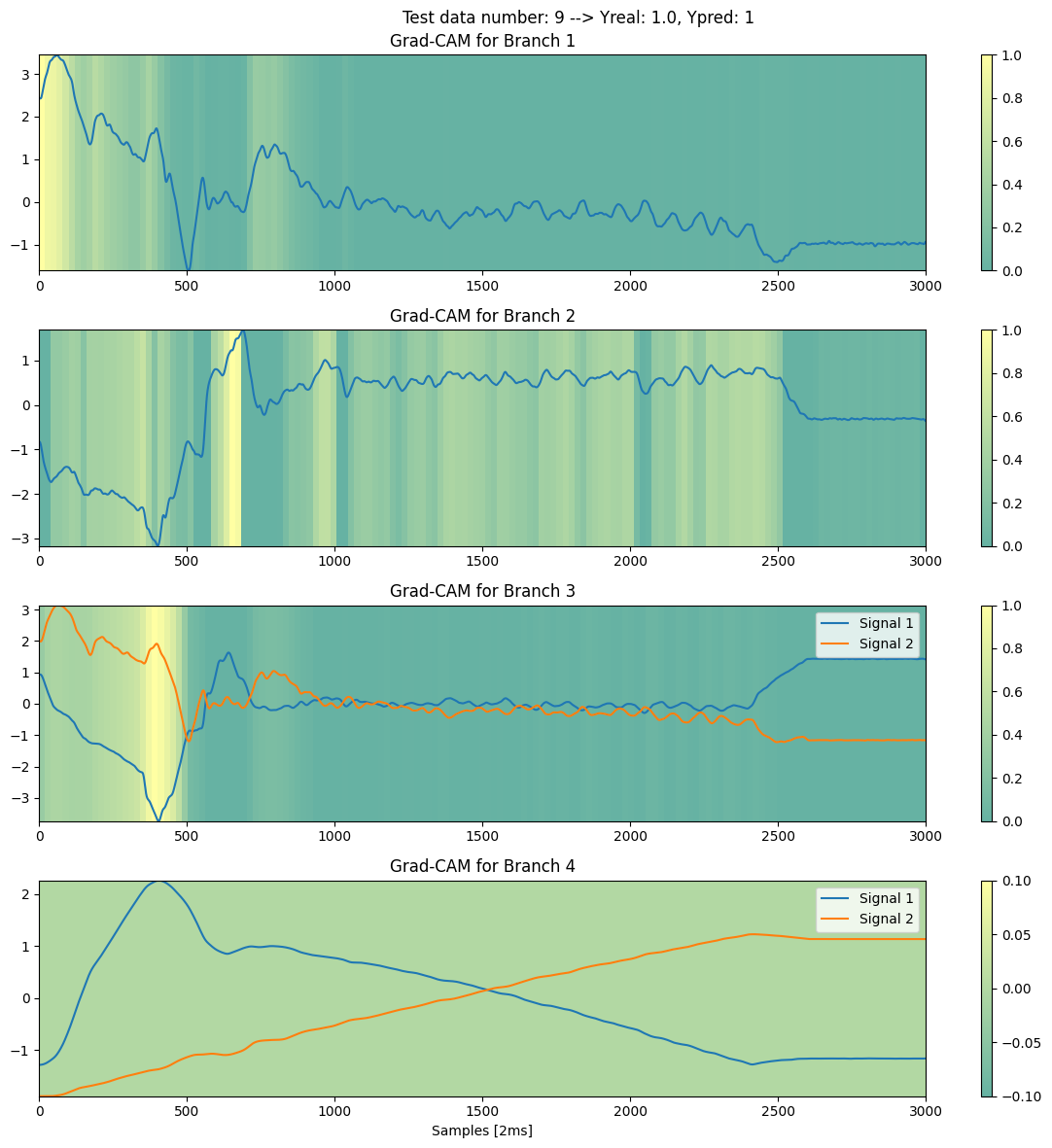}\label{fig:6_1D_GC_LDG_1}}
    \subfloat[Correctly moved and gently released]{\includegraphics[width=0.46\textwidth]{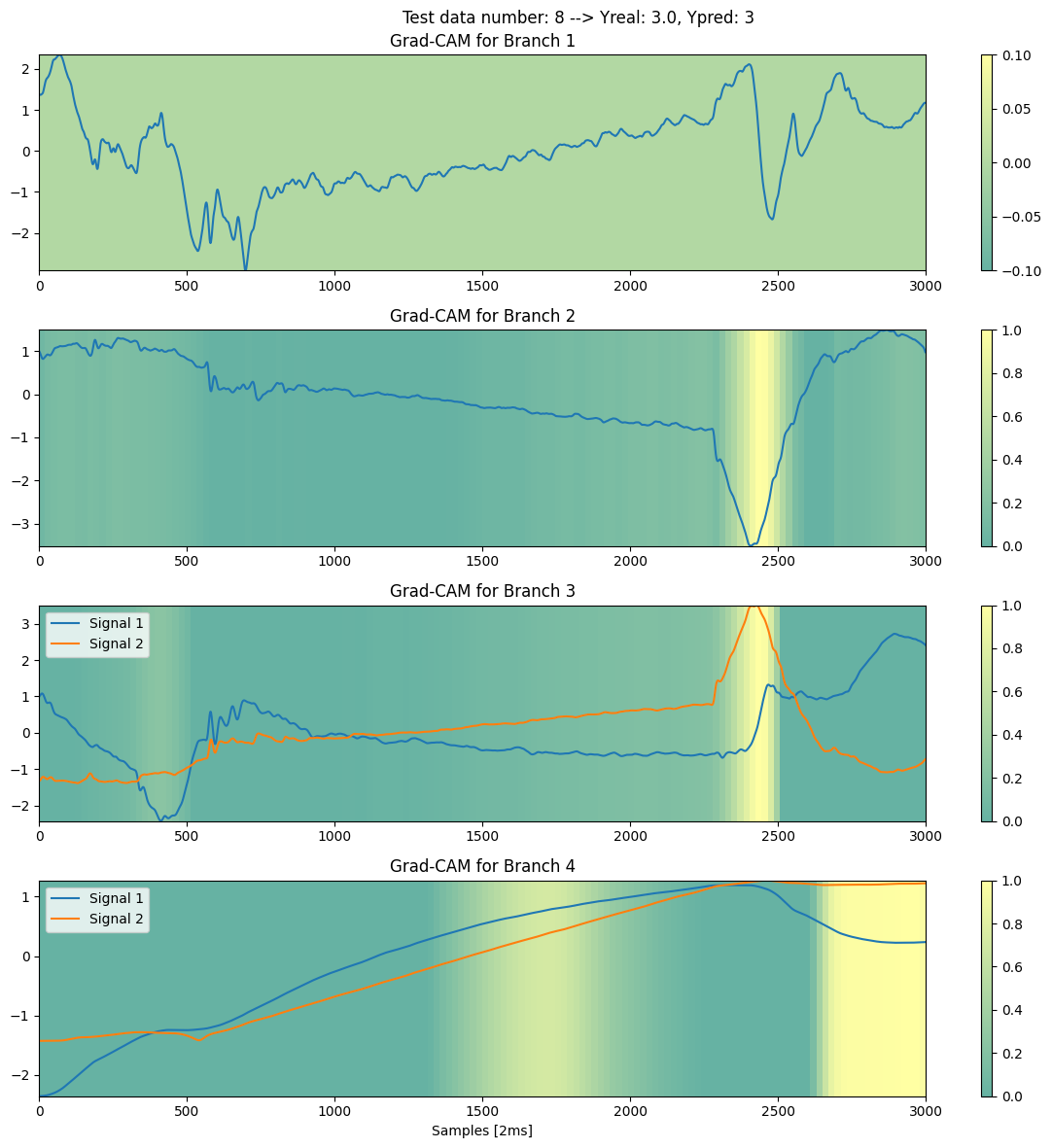}\label{fig:6_1D_GC_LDG_3}}
    \caption{Grad-CAM visualization: (a) The LDG lever is not pulled down with enough force, 
    the model focuses on the initial and mid-part of the signals, where the force is almost constant with ripples, 
    indicating a slow action against resistance, ending without a force spike. (b) The LDG lever is fully moved and released gently, 
    with the model correctly focusing more intensively towards the end of the action to classify success.}
\label{fig:6_1D_LDG_GRADCAM}
\end{figure}

Regarding Knob's action, the initial dataset led to an F1-score of 81\% 
with drastic overfitting due to dataset imbalance since cases 
where the knob was not turned sufficiently were underrepresented. 
To address this, we augmented the dataset by up to half its size, focusing on less frequent cases, 
and employed extensive \texttt{Dropout} and \texttt{MaxPooling}.
This approach yielded a final F1-score of approximately 95\%,  
eliminating the previously observed overfitting. 
However, this significant improvement suggests potential data-snooping, 
reducing trust in the model compared to others.
Figure~\ref{fig:6_1D_LDG_GRADCAM} compare two cases 
of the Landing-Gear lever (LDG) action with insights by 1D-Grad-CAM.
\vspace{-0.3cm}
\vspace{-0.45cm}

\subsection{2D and Hybrid Convolutional Neural Networks}%
\label{subsec:2D_H-cnn}

Since the Wavelet Transform is not applied to the 
$\Delta$Pose data, these data are excluded from 
the 2D-CNN models and the corresponding branches 
in the hybrid models. This omission affects 
the performance of the 2D models. The results 
are summarized in Table~\ref{table:2D_results}. 
Despite this limitation, the approach shows 
feasibility and promising performance, though it 
is not directly comparable to the 1D-CNN models. 
\texttt{Knob} action's classification remains challenging 
due to the unbalanced dataset and lack of augmentation.

These outcomes indicate that while the 2D models perform well overall, 
their reliance solely on the processed 2D scaleograms restricts their 
ability to capture some features present in the raw signals.
Consequently, these results support the investigation of hybrid architectures 
that combine both 1D and 2D processing streams. Such hybrid models are expected 
to better detect fine-grained temporal features while 
retaining the broader context provided by the 2D approach.\\
Three different architectures of Hybrid models are then considered:
\begin{enumerate}
    \item \textbf{All}: All 1D signals are processed in Branch~1, 
    and all 2D scaleograms are  handled in Branch~2.
    \item \textbf{Unit-Measure}: Five branches are used, 
    each processing signals sharing the same unit and with the same format.
    \item \textbf{Specific}: The best 1D-CNN and 2D-CNN models are combined aiming to improve performance and robustness.
\end{enumerate}

\vspace{-0.3cm}
\begin{table}[h]
\centering
\begin{minipage}[t]{0.46\textwidth}
    \centering
    \begin{tabular}{|c|c|c|}
        \hline
        \textbf{Element } & \textbf{F1-Score } & \textbf{Params. } \\ \hline
        Button  & 95.7\% & $\sim$1.1M \\ \hline
        Knob    & 78.5\% & $\sim$1.4M \\ \hline
        Switch  & 93.3\% & $\sim$0.5M \\ \hline
        FLAP    & 89.0\% & $\sim$1.2M \\ \hline
        LDG     & 94.4\% & $\sim$0.7M \\ \hline
        S-Brake & 95.1\% & $\sim$0.6M \\ \hline
    \end{tabular}
    \vspace{0.2cm}
    \caption{2D-CNN Results}\label{table:2D_results}
\end{minipage}
\hfill
\begin{minipage}[t]{0.50\textwidth}
    \centering
    \begin{tabular}{|c|c|c|c|}
        \hline
        \textbf{Element } & \textbf{  All  } & \textbf{Unit-Meas } & \textbf{Specific } \\ \hline
        Button  & 88.2\% & 91.5\% & 96.3\% \\ \hline
        Knob    & 74.8\% & 76.2\% & 80.7\% \\ \hline
        Switch  & 95.4\% & 93.7\% & 97.1\% \\ \hline
        FLAP    & 86.3\% & 91.0\% & 92.5\% \\ \hline
        LDG     & 91.6\% & 90.3\% & 97.8\% \\ \hline
        S-Brake & 94.0\% & 95.8\% & 98.4\% \\ \hline
    \end{tabular}
    \vspace{0.2cm}
    \caption{Hybrid-CNN Results: F1-Score}\label{table:hybrid_results}
\end{minipage}
\end{table}
\vspace{-0.3cm}

\begin{figure}[h]
    \centering
    \includegraphics[width=0.75\textwidth]{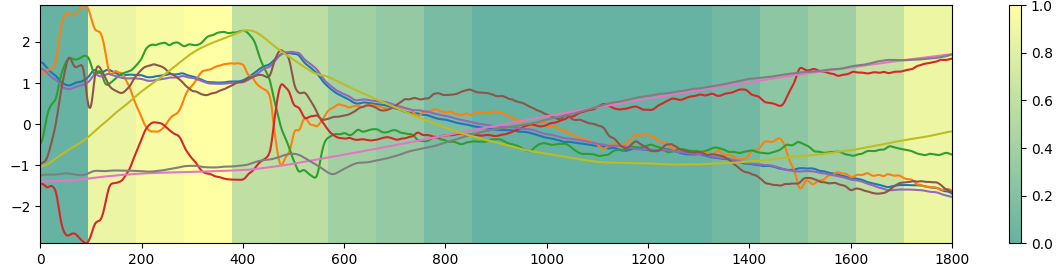}
    \caption{Example of a 1D-GradCAM plot for the Speed Brake action, illustrating the complexity of interpreting multiple uncorrelated inputs.}
    \label{fig:hyb_speed_as1d}
\end{figure}

As anticipated, Specific models yield the highest performance 
by focusing on the most relevant 
signals for each action and leveraging the best-performing models previously 
trained and fine-tuned for specific tasks.
Best performance is observed in actions with balanced data, 
such as Button, Switch, LDG, and S-Brake. 
The Knob action still underperforms because only the 
non-augmented dataset was used. 
Similarly, the FLAP action struggles, as in the 2D-CNN case.
Branches  processing scaleograms seem to adversely affect the overall 
decision process also in the hybrid model.
General models (\textbf{All} and \textbf{Unit-Measure}) perform well with a uniform strategy. 
They require minimal tuning or data selection. 
1D-GradCAM plots provide insights into the decision process, 
verifying that irrelevant features are not considered, 
however, many uncorrelated inputs in a single plot complicate interpretation as in Fig.~\ref{fig:hyb_speed_as1d}.

\vspace{-0.25cm}
\vspace{-0.2cm}

Additionally, since signal lengths vary across actions, 
their scaleograms have different widths, ranging from 800x128 to 3000x128. 
These larger sizes increase memory requirements during training, 
necessitating fewer filters to prevent overflow. Furthermore, 
2D datasets, and consequently Hybrid ones, 
are significantly larger than 1D datasets, 
containing 768 channels compared to 9, 
which is noticeable during data loading.

\vspace{-0.4cm}
\section{Conclusion and Discussion}\label{sec:conclusion}
\vspace{-0.2cm}
These findings confirm the feasibility of the proposed approach, 
which achieves optimal performance on most
action types. The single exception is the knob task, 
where an unbalanced dataset constrains performance;
however, an F1-score exceeding 78\% is still achieved.
The preprocessing pipeline effectively cleans the data 
and mitigates noise from hardware and controller
constraints, while the energy-based transient detection 
algorithm isolates the most informative regions of
each signal.

Among the tested architectures, 1D-CNN models emerge 
as the most efficient option, excelling in accuracy,
simplicity, and training speed.
Although 2D and Hybrid models offer valid alternatives, they come with higher
computational costs but remain trainable in minutes on a standard workstation.

Given the safety-critical nature of the application, the primary goal at this stage, 
is to develop an accurate solution tailored to the specific setup, 
rather than pursuing a broad, general-purpose model. 
The resulting algorithms are therefore well-suited 
to this particular robot configuration and its embedded Force Torque Sensor.
Looking ahead, as the project continues and additional 
data are gathered under diverse conditions from various
simulators and real cockpits, future refinements are expected 
to enable a more generic yet precise wrench or tactile-based
validation algorithm, noting that this subsystem operates 
in parallel with a computer vision-based
validation method, enhancing both
redundancy and overall system robustness.
\vspace{0.2cm}
\\
\texttt{Acknowledgment:} this work is supported by Progetto Accordo 
per l'innovazione: "ARTO - Automatic Robotic for Testing Optimisation", 
code F/350238/01-03/X60.
\vspace{-0.4cm}
%
%
\bibliographystyle{unsrt}
%

\end{document}